# DESIGN OF OUTDOOR AUTONOMOUS MOBILE ROBOT

[1]I-HSI KAO, [2]JIAN-AN SU, [3]JAU-WOEI PERNG

[1,2,3]National Sun Yat-Sen University, Kaohsiung, Taiwan (R.O.C.)
Email: [1]flush0129@gmail.com, [2]40127154@gm.nfu.edu.tw, [3]jwperng@faculty.nsysu.edu.tw
Contact: [123]+886-75253021

*Abstract*—This study presents the design of a six-wheeled outdoor autonomous mobile robot. The main design goal of our robot is to increase its adaptability and flexibility when moving outdoors. This six-wheeled robot platform was equipped with some sensors, such as a global positioning system (GPS), high definition (HD) webcam, light detection and ranging (LiDAR), and rotary encoders. A personal mobile computer and 86Duino ONE microcontroller were used as the algorithm computing platform. In terms of control, the lateral offset and head angle offset of the robot were calculated using a differential GPS or a camera to detect structured and unstructured road boundaries. The lateral offset and head angle offset were fed to a fuzzy controller. The control input was designed by *Q*-learning of the differential speed between the left and right wheels. This made the robot track a reference route so that it could stay in its own lane. 2D LiDAR was also used to measure the relative distance from the front obstacle. The robot would immediately stop to avoid a collision when the distance between the robot and obstacle was less than a specific safety distance. A custom-designed rocker arm gave the robot the ability to climb a low step. Body balance could be maintained by controlling the angle of the rocker arm when the robot changed its pose. The autonomous mobile robot has been used for delivery service on our campus road by integrating the above system functionality.

*Index Terms*: Lane keeping, *Q*-Learning, robotic.

## I. INTRODUCTION

In recent years, with the rapid development of science and technology alongside industrial automation, human beings hope to have a higher quality of life and more comfortable work environments. Robots have gradually replaced manpower, and the invention of many devices has considerably reduced labor costs in various industries. Therefore, this paper will present research results on regional delivery robots.

According to [1], the Starship team is developing a mobile robot that can travel on roads at a speed of 3 km/h. In [2], the TAPAS robot was introduced that can carry 3 to 5 kg loads and can overcome self-navigation problems that are present in park-based environments. Thus, customers can choose from a range of short and precise delivery times and instantly track the robot's position via a mobile app. The robot uses autonomous driving, and safety can be ensured through monitoring.

The six-wheel motion behavior analysis section is slightly different from general differential control [3]. In [4], the six-wheel drive platform is proposed to establish a kinematics model for the vehicle chassis using the Denavit-Hartenberg method. The positive solution to the kinematic model was then used to analyze the pitch of a six-wheel variable-width articulated chassis.

The six-wheeled stage can provide a large contact area and stability compared to a four-wheel system. When the vehicle is designed with a suitable mechanism, it is considered that the stage must be moved to an unknown outdoor site. When a six-wheeled vehicle is moving linearly, it functions like a general four-wheel system. Some changes must be made when turning a corner, because there are multiple sets of drive wheels. In order to minimize the radius of gyration of the car body, this study only uses the middle drive wheelsets during rotation.

In addition to the standard six-wheeled vehicle, there are also deformed mechanism designs. In [5, 6], a drive wheel design was proposed that provides each drive wheel with a rotational degree of freedom. This design allows the vehicle to be more flexible on rough terrain.

In order for robots to move safely and autonomously in complex environments, successful execution of transport and delivery often depends on accurate location information. In [7], the author uses a differential global positioning system to accurately position the robot. The technique requires determining the distance between the transmitting and receiving ends by transmitting the time difference between the satellites and surface receivers around the earth.

In [8], the author used a laser range finder to measure the distance between the front end and obstacles. This allows collisions involving robots to be avoided, and a safe distance from the established target can be maintained. Movement is established using the motion of the robot to push out the position and direction of a certain moment according to the kinematics.

In [9, 10], *Q*-learning and fuzzy controllers are combined and implemented, allowing for control by a human. According to the characteristic information of the sensor grabber, the follower can instantly capture motion in real time.

A six-wheeled platform was developed and is presented in this paper. Three of the wheels include an instant dynamic positioning system, speed and steering control system, and a body balance system. The instant dynamic positioning system ensures the vehicle stays on a planned path. The speed and steering control system determines the control strategy and control commands. The body balance system ensures the vehicle is balanced by controlling the rocker arm in the middle and rear wheels.

The remainder of this paper is organized as follows. The hardware of the three systems and the design concept of the robot are presented in Section II. The algorithm of the three systems is described in Section III. The entire system operation process will be fully introduced. The experimental results and analysis are presented in Section IV. Finally, Section V concludes the paper and suggests topics for future research.

## II. SYSTEM HARDWARE

This paper proposes a set of sensors that can be applied to robot systems. The robot is operated in a closed area, and the local campus is used to verify the self-delivery function. The robot looks like a locker with six wheels, as shown in Figure 1.

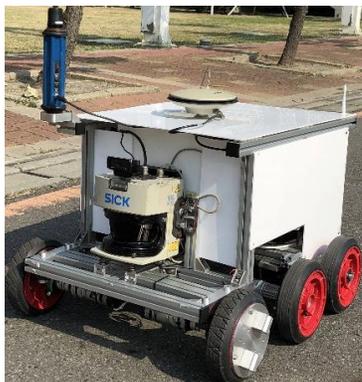

**Figure 1: The designed robot in this work**

The size of the box on the robot is sufficiently large to load several small parcels and will be responsible for transporting groceries and small bags. The cargo carried by the robot should not exceed 8 kg.

The robot is driven by four DC motors that are controlled by the front and middle wheels. When executing a turn, the intermediate wheel is used for differential drive and the front and rear wheels act as freewheels. This design gives the vehicle a small gyration radius. The actuator of the rocker system uses stepper motors and a gear reducer.

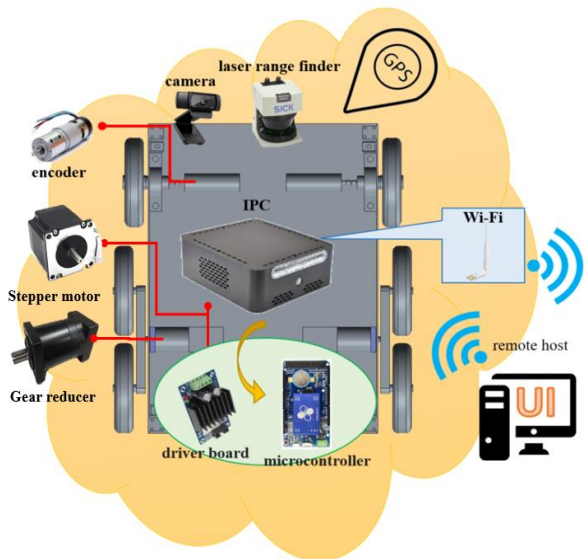

**Figure 2: Hardware structure of the robot**

An industrial computer is used to control the robot. This computer can function in harsh environments, including wet or dusty environments. This design also allows the robot to operate in harsh environments. The computer sends a control signal to the motor with a microcontroller. The LiDAR and camera are installed at the front of the robot for avoidance. The hardware structure of the robot is shown in Figure 2.

This study uses the satellite locator to measure the instantaneous dynamic latitude and longitude coordinates and the heading angle. A reference path is established in advance and is used as the input signal source for horizontal and vertical control of the robot. Two GPS receivers are provided in order to improve the accuracy of the measurement signal. The differential positioning algorithm of the global satellite locator reduces the robot positioning accuracy error to 1 cm ± 1 ppm.

SICK LMS-291-S05 is used to prevent collisions in this study. This equipment uses the time of flight to avoid collisions. Infrared laser light is directed to a magnifying mirror driven by a high-speed rotating motor. The infrared light is refracted by a prism to scatter the infrared beam. Light is reflected off the object and the time of flight is measured as the light pulse moves back and forth through the system, thus the relative position of the object can be determined.

The robot uses an inertial measurement unit (IMU) to complete the balance system. Balance data includes roll, pitch, and yaw. Altitude can be balanced by controlling the middle wheel with the rocker arm system.

## III. SYSTEM PROCESS

### A. *The instant dynamic positioning system*

The current latitude, longitude, and heading angle of the robot is obtained from the satellite positioning data, and the lateral offset and angular deviation of the robot can be calculated. The reference path of the robot is shown in Figure 3.

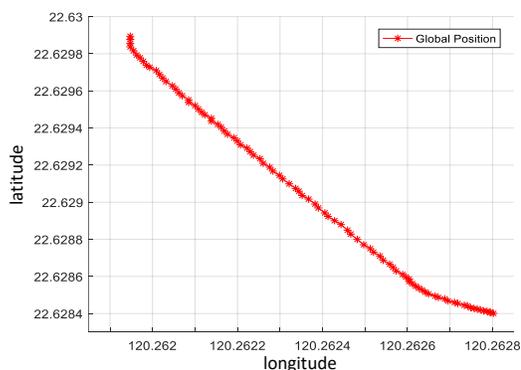

**Figure 3: Reference path**

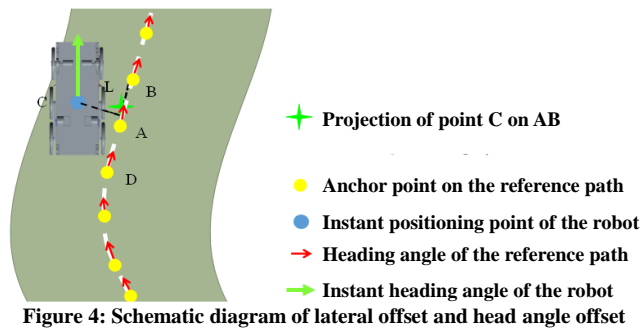

**Figure 4: Schematic diagram of lateral offset and head angle offset**

The path correction method is shown in Figure 4. Bubble sorting is used to find the nearest two reference path points (A and B) from the current robot's anchor point C. Point C is projected on the line connecting A and B to determine L. The distance between L and C is the lateral offset. The closest anchor point A on the established reference path is determined from the current robot position at point C. The angular difference between the heading angle at point A and the current heading angle of the robot can be calculated, and the head angle offset can be obtained.

It is unwise to rely on GPS for positioning. In this study, road boundary detection is used to detect the boundaries of structured and unstructured roads. Lane boundary feature extraction in images is implemented using the Canny algorithm [11], where the Hough Transform is used to convert and detect the road line in two steps: voting and threshold detection. The detected road line will then be used with the heading angle to determine whether it is true to the road boundary. Finally, the focus of the two road boundaries detected is the vanishing point. Lateral offset and angular deviation of the robot can be calculated with this process.

The Canny algorithm has three detection advantages, including low error rate, high positioning accuracy, and minimum response [12]. Low error rate means that the Canny algorithm can mark the edge in the image and reduce noise as much as possible simultaneously. High positioning accuracy means that the marked edge should coincide with the origin edges in the image. Minimum response means that one edge should only be marked once.

The Canny algorithm is completed in four steps, including Gaussian filtering, gradient and angle calculation, non-maximum suppression, and double threshold judgment boundary. Image smoothing technology is used to eliminate low-frequency noise that appears in images. The Gaussian filter function is as follows:

$$G(x,y) = \frac{1}{2\pi\sigma} e^{-\frac{x^2+y^2}{2\sigma^2}}, \quad (1)$$

where $(x, y)$ is the pixel position and $\sigma$ is the standard deviation. Sobel is used to calculate the gradient and angle. The horizontal gradient component ($G_x$) and vertical gradient component ($G_y$) is shown as follows:

$$G(x) = \begin{bmatrix} -1 & 0 & 1 \\ -2 & 0 & 2 \\ -1 & 0 & 1 \end{bmatrix}, \quad (2)$$

$$G(y) = \begin{bmatrix} -1 & -2 & -1 \\ 0 & 0 & 0 \\ 1 & 2 & 1 \end{bmatrix}. \quad (3)$$

The value ($G$) and angle ($\theta$) of the gradient component is calculated as follows:

$$G = \sqrt{G_x^2 + G_y^2}, \quad (4)$$

$$\theta = tan^{-1} \frac{G_y}{G_x}. \quad (5)$$

Through non-maximum suppression, the highest value on the gradient component will be the position of the edge. Finally, a double threshold judgment boundary is used to find the value of the gradient that is between the threshold and determine whether or not the edge is true.

The Hough transform is a feature extraction technique in image processing. The Hough transform was proposed by P. V. C. Hough in 1962 [13]. The original Hough transform was designed to detect straight lines and curves. In 1972, R. Duda and P. Hart improved the Hough transform to distinguish objects of any shape [14]. A Hough transform uses transformations between coordinate spaces. A line or curve with the same shape in space is mapped to a point in another coordinate space and forms a peak. One-to-many mapping is used to map the image space coordinates to the parameter space coordinate. All possible parameter values are then accumulated. Finally, a set of data with the most accumulated times is obtained in the parameter coordinates. This set of data is the shape parameter in the original image space coordinate.

Finally, the final two road boundary lines are obtained from the Hough transform. The two lines are extended forward, and the intersection between the two boundary lines is calculated as the vanishing point. The vanishing point and the center point are used to calculate the lateral offset and head angle offset of the robot for the road boundary line, as shown in Figure 5.

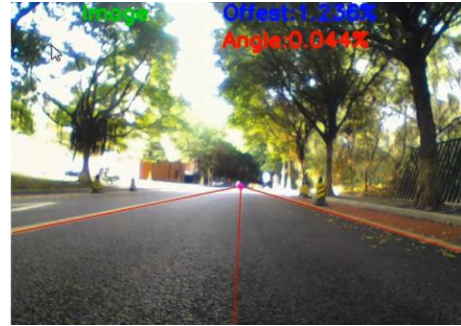

**Figure 5: Lateral offset and head angle offset in the robot image**

*B. Speed and steering control system*

The system presented in this paper relies on the left and right wheel motors to drive independently. The robot uses the principle of sliding steering, where different radii or even in-situ steering are accomplished by changing the speed of the wheels on both sides of the robot. As input to the fuzzy controller, the lateral offset and head deviation angle on the path are calculated by the instant dynamic positioning system and road boundary line detection. The control commands are then fed into motors via Kalman filtering. In this study, *Q*-learning is used to modify the attribution function of the fuzzy controller. The current control strategy is determined by evaluating the current state of the environment. The appropriate action is chosen by allowing the controller to learn the appropriate control strategy on its own. This technology can improve the performance of the controller and allows the delivery robot run safely on the road.

Fuzzy theory was proposed by L. A. Zadeh in 1965 [15]. Since its introduction, fuzzy theory has been widely used in industry, factory automation, and in other fields. The fuzzy algorithm can establish a clear physical-mathematical model under a controlled system with uncertainty. This design allows the controller to reduce its dependence on mathematical models. The fuzzy controller has a complete design flow [16]. The overall flow of the fuzzy control algorithm in this study is shown in Figure 6. The robot control system designed in this thesis adopts dual input and dual output architecture to control motion based on fuzzy theory. First, the robot's instantaneous latitude, longitude, and heading angle information are gathered from the satellite locator. The robot will track the latitude, longitude, and head angle of the ideal planned path and lane detection during imaging. The lateral offset and angular offset of the robot are calculated from the GPS signal and the image. The lateral offset and angular offset are the inputs to the fuzzy controller. After applying the fuzzy inference method, the left and right wheel controls can be dynamically output to control the robot's motion.

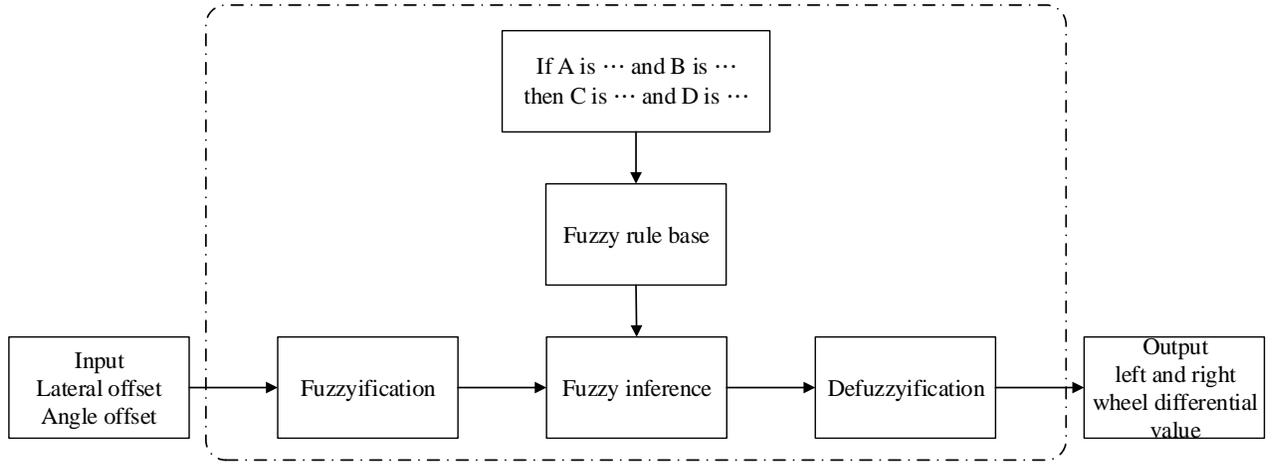

Figure 6: Fuzzy algorithm flowchart

The fuzzy set is defined as follows:

$$X = \{Left, Moderate, Right\}$$
$$\theta = \{Left, Moderate, Right\}, \quad (7)$$
$$V = \{Low, Moderate, High\}$$

where $X$ is the lateral offset, $\theta$ is the angle offset, and $V$ is the differential wheel speed. The input and output variables are defined using triangular and trapezoidal attribution functions. The attribution function of the lateral offset is as follows:

$$Left = [X_{L0}, X_{L2}] = [-\infty, 0]$$
$$Moderate = [X_{L1}, X_{L3}] = [-10, 10], \quad (8)$$
$$Right = [X_{L1}, X_{L4}] = [0, \infty]$$

where $X_{L0}$ is the left end of the attribution function in the left side, $X_{L1}$ is the left end of the attribution function in the moderate side, $X_{L2}$ is the right end of the attribution function in the left side and the left end of the attribution function in the right side, $X_{L3}$ is the right end of the attribution function in the moderate side, and $X_{L4}$ is the right end of the attribution function in the right side. The attribution function of the angle offset is as follows:

$$Left = [\theta_{L0}, \theta_{L2}] = [-\infty, 0]$$
$$Moderate = [\theta_{L1}, \theta_{L3}] = [-1, 1], \quad (9)$$
$$Right = [\theta_{L1}, \theta_{L4}] = [0, \infty]$$

where $\theta_{L0}$ is the left end of the attribution function in the left side, $\theta_{L1}$ is the left end of the attribution function in the moderate side, $\theta_{L2}$ is the right end of the attribution function in the left side and the left end of the attribution function in the right side, $\theta_{L3}$ is the right end of the attribution function in the moderate side, and $\theta_{L4}$ is the right end of the attribution function in the right side. The attribution function of the differential wheel speed is as follows:

$$Low = [V_{L0}, V_{L2}] = [-\infty, 50]$$
$$Mid = [V_{L1}, V_{L3}] = [0, 100]. \quad (10)$$
$$High = [V_{L1}, V_{L4}] = [50, \infty]$$

where $V_{L0}$ is the left end of the attribution function in the left side, $V_{L1}$ is the left end of the attribution function in the moderate side, $V_{L2}$ is the right end of the attribution function in the left side and the left end of the attribution function in the right side, $V_{L3}$ is the right end of the attribution function in the moderate side, and $V_{L4}$ is the right end of the attribution function in the right side.

The next step involves building a fuzzy rule base. The output of the fuzzy controller is calculated using the input lateral offset and head angle offset. The fuzzy rule base is listed in Tables 1 and 2. The term "fuzzification" refers to making the input variables fuzzy. While $< X_{L1}$, the value of the attribution function in the left side is shown as follows:

$$\mu X_0 = 1. \quad (11)$$

While $X_{L1} < X < X_{L2}$, the value of the attribution function in the left side and moderate side are shown as follows:

$$\mu X_0 = (X_{L2} - X)/(X_{L2} - X_{L1}), \quad (12)$$
$$\mu X_1 = (X - X_{L1})/(X_{L2} - X_{L1}). \quad (13)$$

While $X_{L2} < X < X_{L3}$, the value of the attribution function in the moderate side and the right side are shown as follows:

$$\mu X_1 = (X_{L3} - X)/(X_{L3} - X_{L2}), \quad (14)$$
$$\mu X_2 = (X - X_{L2})/(X_{L3} - X_{L2}). \quad (15)$$

While $X_{L3} < X$, the value of the attribution function in the right side is shown as follows:

$$\mu X_2 = 1. \quad (16)$$

Fuzzification of the heading angle offset is the same as the lateral offset.

Max-min composition is used for fuzzy inference in this study. This model is developed by first taking the fuzzy set cooperative intersection operation to the minimum value. The fuzzy set union operation is then used to take the maximum value. Finally, the fuzzy output is obtained. The main operation

## Table 1
### Fuzzy rule base of the left wheel

|  |  | Lateral Offset | | |
|---|---|---|---|---|
|  |  | Left | Moderate | Right |
| Heading Angle Offset | Left | High Speed | High Speed | Mid Speed |
|  | Moderate | High Speed | Mid Speed | Low Speed |
|  | Right | Mid Speed | Low Speed | Low Speed |

## Table 2
### Fuzzy rule base of the right wheel

|  |  | Lateral Offset | | |
|---|---|---|---|---|
|  |  | Left | Moderate | Right |
| Heading Angle Offset | Left | Low Speed | Low Speed | Mid Speed |
|  | Moderate | Low Speed | Mid Speed | High Speed |
|  | Right | Mid Speed | High Speed | High Speed |

formula is as follows:

$$\mu_{L,R\_i} = Max\big(Min(\mu X_i, \mu \theta_i), \mu_{L,R\_i}\big), \quad (17)$$

where $\mu_{L,R\_i}$ is the value of the wheel attribution functions.

Defuzzification is the last design step of the fuzzy controller. The output obtained after fuzzy inference is the amount of blur, so it is necessary to obtain the actual output through defuzzification. The area method is used as the method of defuzzification in this study. The output calculation method for the left driven wheel is as follows:

$$CL_1 = (V_{L3} - V_{L2})/\big(1 - (1 - \mu_{L\_2})^2\big), \quad (18)$$

$$CL_2 = CL_1 \cdot V_{L3}, \quad (19)$$

$$BL_1 = (V_{L3} - V_{L1})/2 \cdot \big(1 - (1 - \mu_{L\_1})^2\big), \quad (20)$$

$$BL_2 = B_1 \cdot V_{L2}, \quad (21)$$

$$AL_1 = (V_{L2} - V_{L1})/\big(1 - (1 - \mu_{L\_0})^2\big), \quad (22)$$

$$AL_2 = A_1 \cdot V_{L1}, \quad (23)$$

$$V_{t\_left} = (AL_2 + BL_2 + CL_2)/(AL_1 + BL_1 + CL_1), \quad (24)$$

where $CL_1$ is the trapezoidal area of the attribution function at high speed, $CL_2$ is the trapezoidal area of the attribution function at high speed multiplied by the left end of the attribution function at moderate speed, $BL_1$ is the trapezoidal area of the attribution function at moderate speed, $BL_2$ is the trapezoidal area of the attribution function at moderate speed multiplied by the right end of the attribution function at low speed, $AL_1$ is the trapezoidal area of the attribution function at low speed, $AL_2$ is the trapezoidal area of the attribution function at low speed multiplied by the left end of the attribution function at moderate speed, and $V_{t\_left}$ is the speed control command. Defuzzification of the right wheel is similar to the left wheel.

The relationship between the two input and the double output variables is obtained from equations (18) to (24). The fuzzy control surface of the left wheel speed is shown in Figure 7. One can see from the surface that both input variables need to be normalized to a range of -1 to 1 in order to limit the variation between the input variable and the output variable to the surface of the intermediate block. The control block diagram is shown in Figure 8.

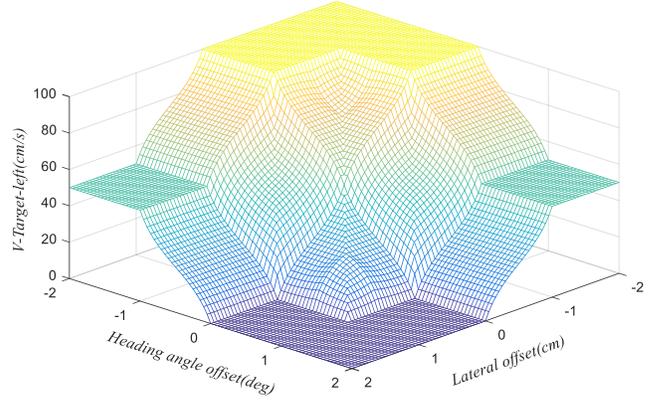

**Figure 7: The fuzzy control surface of left wheel speed**

$Q$-Learning is a reinforcement learning algorithm [17]. During operation, the system assigns a $Q$-table to each state for each $Q$-function. After accumulating learning experience for each action, the obtained $Q$-function is:

$$Q^+ = \max_A Q(S_{k+1}, A_{k+1}), \quad (25)$$

where $S_k$ is the state vector, $A_k$ is the motion vector, and $Q(S_k, A_k)$ is the action function. After many studies, an ideal value $Q^+$ will be reached. $Q^+$ is used to choose the $Q$ maximum value of the next state, i.e., each time the selected value is the largest. According to the feedback value obtained after the action, the update equation for reaching the new state and obtaining the expected reward in the future is as follows:

$$Q(S_k, A_k) \leftarrow Q(S_k, A_k) + \alpha \Big[r_{k+1} + \gamma \max_A Q(S_{k+1}, A_{k+1}) - Q(S_k, A_k)\Big], (26)$$

where α is the learning rate, $r_{k+1}$ is the immediate feedback value according to $A_k$, $\max_A Q(S_{k+1}, A_{k+1})$ is the expected maximum value that can be obtained in state $S_k$ according to the experience accumulated so far, and γ is the decay rate. If γ is smaller, it means that the future value expected from the current time is valued less. Therefore, each $Q$-learning step will correspond to a $Q$-table that indicates how each state corresponds to the value of a given action.

The $Q$-learning algorithm modifies the fuzzy attribution function. Each $Q$-function is recorded and the established state

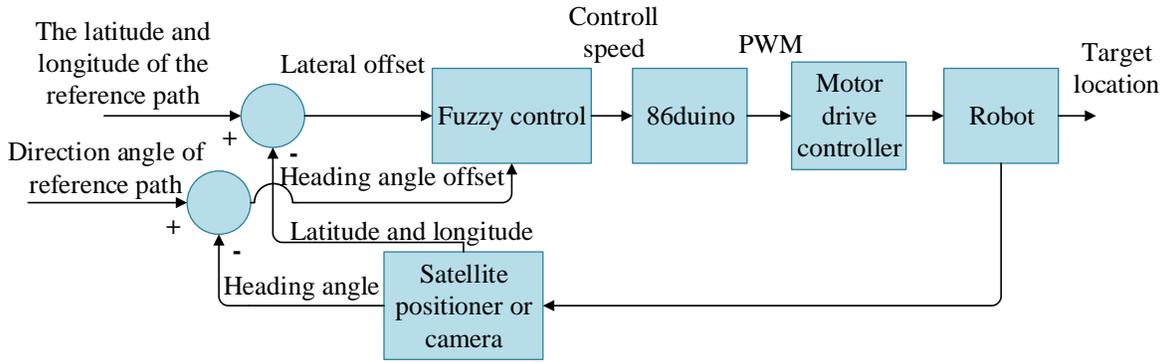
Figure 8: Control block diagram

and the corresponding action $Q$-table are corrected, thus allowing the robot to learn interactively from its environment. Since robots may encounter scenes in an unknown environment, robots learn their strategies by making judgments based on the $Q$-function in various states while interacting with the environment, and correcting errors in real world environments.

In this study, the input attribution functions are distance and angle, and there are four parameters that can be corrected, as shown in Figure 9. In the adjustment process, it is found that the distance adjustment will affect the speed and stability of the robot. Therefore, this experiment will explore robot motion while adjusting distance.

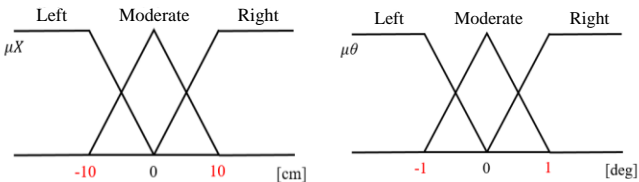
Figure 9: Four parameters adjusted by Q-Learning

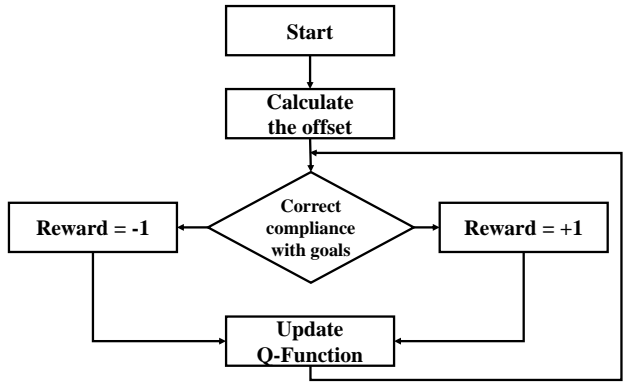
Figure 10: Reward and punishment process

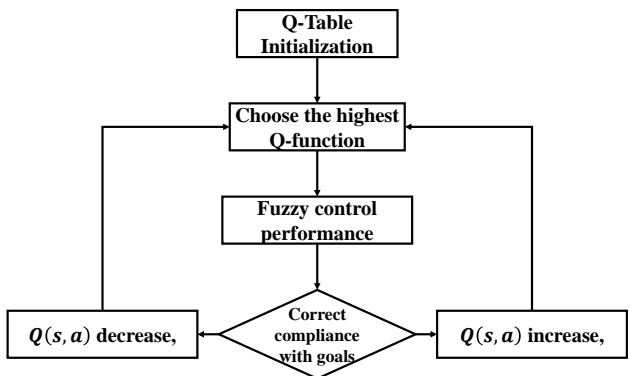
Figure 11: Flowchart of Fuzzy control combined with $Q$-learning

After many corrections to the attribution function, the robot can obtain ideal control. During the learning process, the value of the action corresponding to the state is classified as good or bad, and the value of this action will be appropriately rewarded by comparing distance error changes between the current state and the previous state. If the adjustment effect is undesirable, a penalty will be applied to the value of the state action in the $Q$-table. The use of bonus and deduction points as the strength of the feedback signal is shown in Figure 10. The combination of the ideal fuzzy control and $Q$-learning algorithm is shown in Figure 11.

Anti-collision is also an important issue when controlling robots. In this work, LiDAR is used as a sensor for detecting obstacles at the front of the robot. Cluster analysis is based on the relationship between multiple objects. When the similarity within a group is high, and the difference from other groups is very large, a representative group can be obtained. In this work, density-based spatial clustering of applications with noise (DBSCAN) is used as an algorithm for clustering LiDAR data [18]. DBSCAN is a density-based clustering algorithm. Compared with a general clustering algorithm, it is necessary to specify the number of classifications, such as is done in the K-means algorithm [19]. DBSCAN finds many clusters that cannot be found with K-means.

*C. Body balance system*

This system uses a 57 stepper motor with a drive and a 1:216 reducer as a key component. By using the motor to drive the designed rocker system, the middle wheel can be raised to achieve balance and stability.

The robot can adapt to some terrain in an outdoor environment by correcting its posture [20]. Figure 12 shows the process used to balance the body, drive the stepping motor, and drive the rocker arm through the reducer to lift the middle wheel.

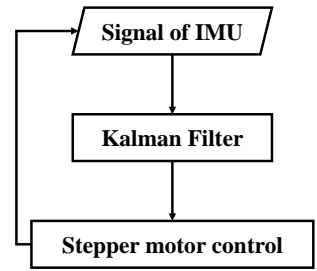
Figure 12: Flowchart of the body balance system

The signal from the IMU is smoothed with a Kalman Filter.

A Kalman Filter is a high-efficiency linear filtering recursive formula [21]. Each time a new measurement is added, a new state value can be estimated based on the previous state data that has been calculated with the system. When using the Kalman filter, it is necessary to establish a discrete state space model. The state model and measurement model of the filter in a discrete nonlinear system are as follows:

$$x_{k+1} = f(x_k) + w_k, \quad (27)$$

$$z_k = h(x_k) + v_k, \quad (28)$$

where $x_k$ is the state vector, $f(x_k)$ is the system model, $w_k$ is the system noise, $z_k$ is the measurement vector, $h(x_k)$ is the measurement model, and $v_k$ is the measurement noise.

In the discrete nonlinear system state model and measurement model, the average value of the noise is 0, i.e., white noise. The Kalman filter includes initialization, state prediction, state correction, and update processes.

During initialization, it is necessary to give the initial state $x(0|0)$ of the system. The same must be given the initial covariation matrix $p(0|0)$, the systematic error covariation matrix $Q$, and the measurement error covariation matrix $R$. During state prediction, the previous state and the covariation matrix are substituted into the following equation to calculate the current state prediction value and the covariation matrix:

$$x(k+1|k) = F(k)x(k|k) + G(k)u(k), \quad (29)$$

$$p(k+1|k) = F(k)p(k|k)F^T(k) + Q(k), \quad (30)$$

where $F(k)$ is the transformation, $G(k)$ is the input gain, and $u(k)$ is the input. During state correction and update, the corresponding covariation matrix $S$, gain matrix $K$, and innovation matrix $N$ of the Kalman filter are calculated through $x(k+1|k)$ and $p(k+1|k)$ as follows:

$$S(k+1) = H(k+1)p(k+1|k)H^T(k+1) + R(k+1), \quad (31)$$

$$K(k+1) = p(k+1|k)H^T(k+1)/S(k+1) \quad (32)$$

$$N(k+1) = z(k+1) - x(k+1|k) \quad (33)$$

The state update, correction value $x(k+1|k+1)$, and covariation matrix $p(k+1|k+1)$ at this time can be obtained from the above formula.

This system allows the robot to carry a package in the loading box and maintain horizontal balance and stability, thus ensuring the goods are not damaged. The front wheel also has a hanging function, as shown in Figure 13.

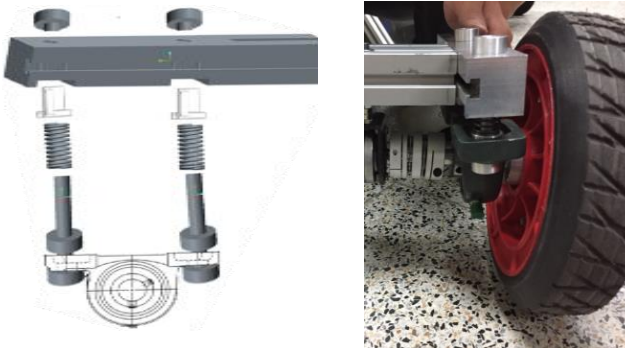

**Figure 13: Front wheel suspension**

## IV. EXPERIMENTAL RESULTS

After completing the vehicle structure and installing the sensing equipment, the robot must pass a field test to prove the functional feasibility of the vehicle. The robot was mainly tested in the surrounding environment of the school campus. The environment includes one-way roads, sidewalks, slopes, and indoor areas. Through these different field tests, the stability, control actions, and perceptions are verified in order to demonstrate the application of this robot.

### A. The instant dynamic positioning system

The GPS positioning experiment is very simple. Precise positioning of the robot can be determined by disassembling the RTK-GPS package. Then, by collecting the GPS positioning points of the robot, the walking trajectory of the robot can be drawn on a Google map, as shown in Figure 14.

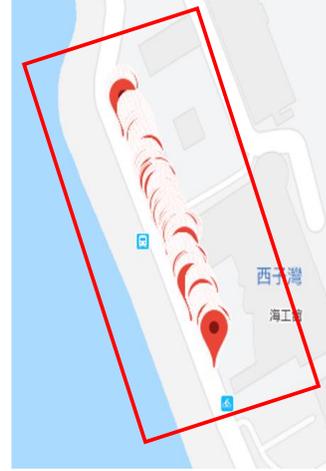

**Figure 14. The path of the robot**

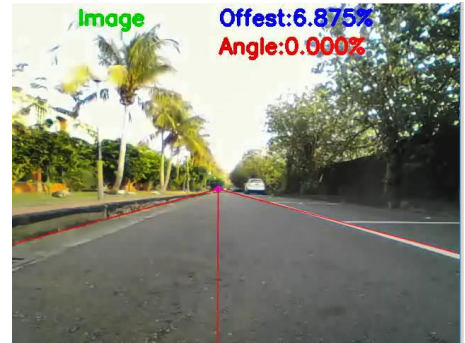

**Figure 15: Single side structure detection results**

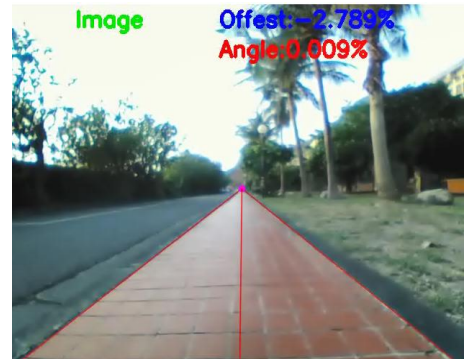

**Figure 16: Sidewalk detection results**

The road boundary detection experimental results are shown in Figures 15-17. The system performs road boundary detection with the front camera. When the GPS signal is blocked or offset, the image road detection system can be switched instantly for control. Two mode switching mechanisms are used to make the six-wheel robot control more stable. Road boundary detection is also performed for different experimental scenarios.

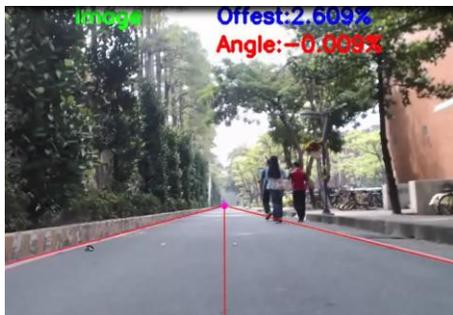

**Figure 17: One-way road boundary detection results**

The control input required by the robot in each path can be obtained from the road boundary lines detected in the above scenarios.

*B. Speed and steering control system*

Motion control for this system is linked to autonomous learning. The attribution function generated by the expert's subjective identification method is compared with the effect of using $Q$-learning to correct motion, and the results are shown in Figure 18. The path error controlled only by fuzzy logic is relatively larger than the method obtained from adding Q-learning. The appropriate action is selected in the environment based on changes in the $Q$-learning state, and feedback is used to judge whether the current action is good or bad. Thus the robot can pursue the target and control its motion. The experimental results show that Q-learning can improve motion control.

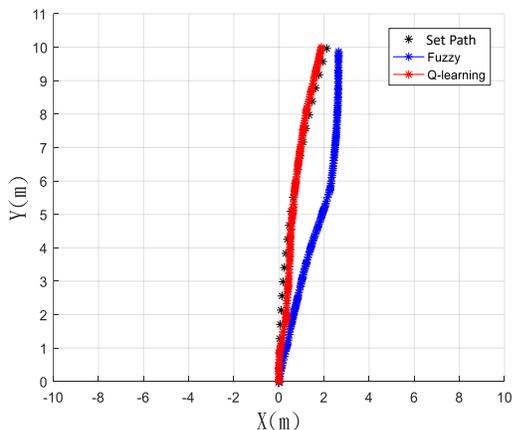

**Figure 17: Comparison of the path between the methods**

Data from the laser range finder were used for collision prevention. When the robot encounters an object as shown in Figure 18, it will stop moving within the set safety range according to the distance from the obstacle, and the speed will drop to 0 mm/s. When the obstacle is removed, the robot begins to accelerate and continue moving. The speed of the robot as it encounters an obstacle is shown in Figure 19.

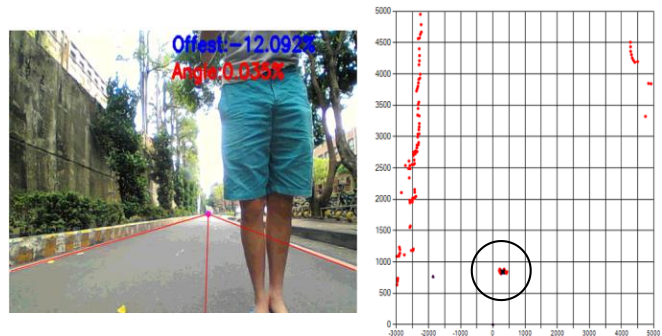

**Figure 18: Obstacle detection result using LiDAR**

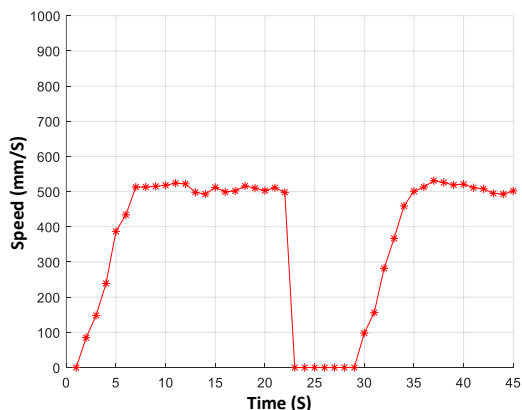

**Figure 19: Robot speed over time as the robot encounters an obstacle**

*C. Body balance system*

When the altitude of the robot changes, the IMU returns its own altitude message and brings it into the Kalman tracking drive stepper motor for balance adjustment. This system ensures that the robot's posture balance and the package in the loading box remain stable and free from slipping.

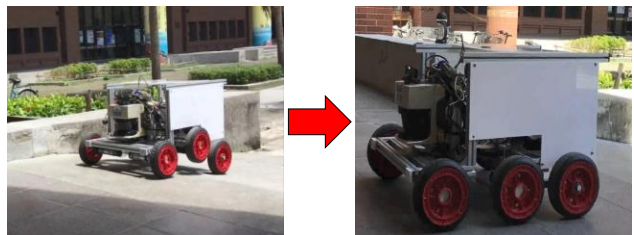

**Figure 20: Body balance experiment diagram**

Results from a test where the robot moved on a slope are shown in Figure 20. When a slope is encountered, the balancing system will maintain balance. The difference between the presence or absence of a balanced system is shown in Figure 21.

After comparing the results, there is no correction of the body rocker arm system, and the altitude angle changes by a dozen degrees. Thus, the feasibility of the system is verified.

In order to ensure a high degree of flexibility in outdoor terrain, the robot was tentatively configured set to climb up to 8 cm. At the beginning of the current climbing step, the motor simultaneously drives the rocker arm to raise the middle wheel and balance the body, and the front wheel continues to move forward. When the intermediate drive wheel is in contact with the ground, the rear freewheel will contact the plane to complete the terrain across the small steps. The experimental process of spanning small steps is shown in Figure 22.

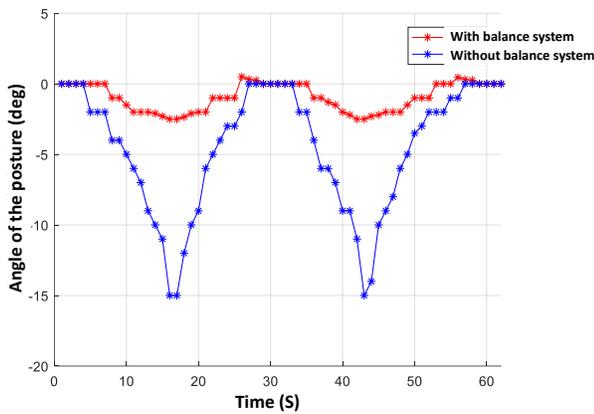

Figure 21: Experimental results with or without a balanced system

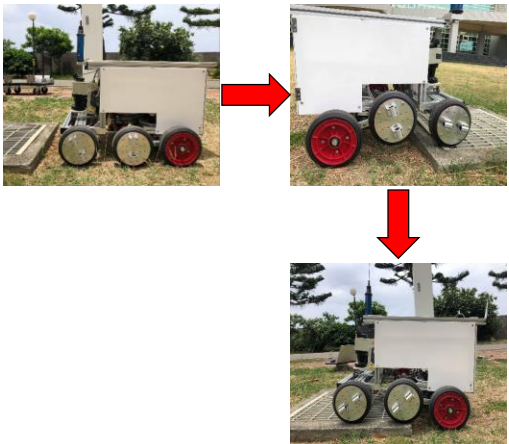

Figure 22: The experimental process of spanning the small steps

## V. CONCLUSION

This work combines platform design and assembly of multiple sensors to develop an outdoor delivery robot. The system uses global satellite positioning, visual road line detection, front obstacle detection, and safely features in an interactive environment according to a planned path, and it completes distribution tasks within a specified area.

The six-wheel structure platform has high mobility and stability, and the rocker arm system is regulated by feedback information from the sensor, which is beneficial for traversing road surfaces and maintaining balance. The lateral and angular offsets are calculated using GPS positioning data and road line detection. The obtained path information is adjusted by the fuzzy controller and $Q$-learning architecture to implement a self-control strategy when the road surface or load changes. The obstacle detection system uses a density-based clustering algorithm to detect objects that can block the robot and effectively avoid collisions. The aforementioned system combines positioning information, environmental sensing, obstacle avoidance technology, and a body suspension structure in a six-wheeled robot.

In the future, we will add anti-theft and identification system functions that allow only the recipient to open the cargo box. At this stage, robots cannot think about problems like humans and cannot handle problems properly. They can only operate through intervention. In terms of performance, I also hope to increase the load weight, speed, and use robots to transport orders over short distances. We hope to gradually implement more complex shipping services in the future.